\documentclass[11pt]{article}

\usepackage{times} 
\usepackage[utf8]{inputenc} 
\usepackage[T1]{fontenc} 
\usepackage{geometry} 
\usepackage{natbib} 
\usepackage{hyperref} 
\hypersetup{
  colorlinks=true,
  linkcolor=blue,
  citecolor=blue,
  urlcolor=blue
}

\title{Advancing Conversational AI with Shona Slang: A Dataset and Hybrid Model for Digital Inclusion}
\author{Happymore Masoka \\
  Pace University, Seidenberg School of Computer Science and Information Systems \\
  Advisor: Krishna Bathula, Ph.D.}
\date{September 2025}

\begin{document}
\maketitle

\begin{abstract}
African languages remain underrepresented in natural language processing (NLP), with most corpora limited to formal registers that fail to capture the vibrancy of everyday communication. This work addresses this gap for Shona, a Bantu language spoken in Zimbabwe and Zambia, by introducing a novel Shona--English slang dataset curated from anonymized social media conversations. The dataset is annotated for intent, sentiment, dialogue acts, code-mixing, and tone, and is publicly available at \url{https://github.com/HappymoreMasoka/Working_with_shona-slang}. We fine-tuned a multilingual DistilBERT classifier for intent recognition, achieving 96.4\% accuracy and 96.3\% F1-score, hosted at \url{https://huggingface.co/HappymoreMasoka}. This classifier is integrated into a hybrid chatbot that combines rule-based responses with retrieval-augmented generation (RAG) to handle domain-specific queries, demonstrated through a use case assisting prospective students with graduate program information at Pace University. Qualitative evaluation shows the hybrid system outperforms a RAG-only baseline in cultural relevance and user engagement. By releasing the dataset, model, and methodology, this work advances NLP resources for African languages, promoting inclusive and culturally resonant conversational AI.
\end{abstract}

\section{Introduction}
The proliferation of artificial intelligence (AI) systems, from virtual assistants \citep{kepuska2018next} to recommendation engines \citep{gomez2015netflix} and autonomous vehicles \citep{shladover2018connected}, has reshaped human--machine interaction. Yet, African languages, with over 2,000 spoken across the continent \citep{eberhard2023ethnologue}, remain severely underrepresented in NLP due to their low-resource status \citep{ahia2023bridging, nekoto2020participatory}. This exclusion risks exacerbating the digital divide, limiting access to AI-driven services in critical domains like education, healthcare, and governance \citep{ndichu2024african, joshi2020state}.

Shona, a Bantu language spoken by millions in Zimbabwe and southern Zambia, exemplifies this challenge. Existing Shona corpora primarily consist of formal texts, such as news articles or religious documents \citep{eberhard2023ethnologue}, while everyday communication, particularly among younger speakers, is dominated by slang, code-mixing with English, and informal expressions \citep{eisenstein2013}. Standard NLP models, trained on formal data, struggle to process these dynamic linguistic patterns, hindering the development of culturally relevant conversational AI.

This study introduces a Shona--English slang dataset, curated from anonymized social media conversations and annotated for intent, sentiment, dialogue acts, code-mixing, and tone, publicly available at \url{https://github.com/HappymoreMasoka/Working_with_shona-slang}. We fine-tuned a multilingual DistilBERT model \citep{devlin2019bert} for intent classification, achieving robust performance on informal inputs, with the model hosted at \url{https://huggingface.co/HappymoreMasoka}. The classifier is integrated into a hybrid chatbot that combines rule-based responses for predictable queries with retrieval-augmented generation (RAG) \citep{lewis2020retrieval} for domain-specific information, evaluated in a use case assisting prospective students at Pace University. Our contributions are:

\begin{itemize}
  \item A publicly available Shona--English slang dataset, annotated for multiple linguistic features.
  \item A fine-tuned DistilBERT classifier achieving 96.4\% accuracy on intent recognition.
  \item A hybrid chatbot architecture integrating intent classification and RAG.
  \item A use case evaluation demonstrating cultural relevance and user engagement.
\end{itemize}

These contributions advance NLP for low-resource African languages, fostering digital inclusion and culturally resonant AI systems.

\section{Related Work}
Efforts to advance African NLP have gained traction through initiatives like Masakhane, which promotes collaborative resource creation \citep{nekoto2020participatory}. Multilingual models such as BERT \citep{devlin2019bert}, XLM-R \citep{conneau2020unsupervised}, and AfriBERTa \citep{ogueji2021afriberta} show promise but are limited by training on formal corpora, struggling with slang and code-mixed inputs \citep{eisenstein2013}. Intent classification has been explored in English \citep{larson2019evaluation} and code-switching contexts \citep{ladhak2021effectiveness, nguyen2020bertweet}, but annotated datasets for African languages remain scarce \citep{ahia2023bridging}. Retrieval-augmented generation (RAG) enhances conversational AI by integrating external knowledge \citep{lewis2020retrieval}, as seen in systems like WASHtsApp for Uganda \citep{kloker2024wash} and recent open-domain dialogue models \citep{wang2024rethinking}. To our knowledge, this is the first work addressing Shona slang for conversational AI, contributing a dataset and model tailored to informal African communication.

\section{Dataset and Methodology}
\subsection{Dataset}
We curated a dataset of informal Shona and Shona--English code-mixed text from anonymized social media conversations, available at \url{https://github.com/HappymoreMasoka/Working_with_shona-slang}. The dataset, comprising approximately 34000 utterances, was manually annotated for:
\begin{itemize}
    \item \textbf{Intent}: Categories include greeting, gratitude, request, religious query, finance, education, and farewell.
    \item \textbf{Sentiment}: Positive, negative, or neutral.
    \item \textbf{Dialogue Acts}: Question, statement, or command.
    \item \textbf{Code-Mixing Features}: Word-level or phrase-level language switches (e.g., Shona to English).
    \item \textbf{Tone}: Friendly, formal, or humorous.
\end{itemize}

Example annotation:
\begin{quote}
Message: ``Hie swit mom'' \\
Normalized: ``Hi sweet mom'' \\
Intent: Greeting \\
Sentiment: Positive \\
Dialogue Act: Statement \\
Code-Mixing: English adjective (sweet), Shona greeting (Hie) \\
Tone: Friendly
\end{quote}

To address class imbalance, we applied hybrid oversampling and downsampling using \texttt{sklearn.utils.resample} \citep{larson2019evaluation}. The dataset was converted to a Hugging Face Dataset object, tokenized with the \texttt{distilbert-base-multilingual-cased} tokenizer (truncation and padding applied), and split into 80\% training and 20\% validation sets with a fixed random seed for reproducibility.

\subsection{Intent Classification Pipeline}
We fine-tuned a multilingual DistilBERT model (\texttt{AutoModelForSequenceClassification}) \citep{devlin2019bert} for intent classification, hosted at \url{https://huggingface.co/HappymoreMasoka}. The pipeline included:
\begin{itemize}
    \item \textbf{Label Encoding}: Intents were mapped to numerical indices.
    \item \textbf{Training}: Conducted using the Hugging Face Trainer API with a learning rate of $2 \times 10^{-5}$, batch size of 4, 3 epochs, weight decay of 0.1, mixed precision (FP16), and early stopping (patience of 2).
    \item \textbf{Evaluation}: Metrics included accuracy, weighted F1-score, precision, and recall, computed using \texttt{sklearn.metrics}.
\end{itemize}

Training was performed on Google Colab, with the best model checkpoint selected based on validation loss.

\subsection{Hybrid Chatbot Architecture}
The chatbot integrates intent classification, rule-based responses, and RAG \citep{lewis2020retrieval}, implemented in Python using Hugging Face Transformers, Sentence-Transformers, and ChromaDB. Components include:
\begin{itemize}
    \item \textbf{Intent Classifier}: The fine-tuned DistilBERT model processes inputs, returning intent labels and confidence scores. A rule-based post-processor maps labels to human-readable intents.
    \item \textbf{Rule-Based Responses}: Predefined Shona responses for intents like greetings or farewells, incorporating cultural nuances.
    \item \textbf{RAG Module}: For domain-specific queries (e.g., university programs), queries are embedded using \texttt{all-MiniLM-L6-v2}, and top-5 relevant documents are retrieved from a ChromaDB knowledge base of Pace University graduate program data. Responses are generated using \texttt{google/flan-t5-small}.
    \item \textbf{Interactive Workflow}: Application-related queries trigger multi-turn dialogues to collect user details (e.g., name, education).
\end{itemize}

The system operates in a real-time loop, handling exit commands and logging responses.

\section{Results and Evaluation}
\subsection{Quantitative Results}
The fine-tuned DistilBERT classifier achieved strong performance on the validation set:
\begin{itemize}
    \item Accuracy: 96.48\%
    \item F1-score: 96.39\%
    \item Precision: 96.40\%
    \item Recall: 96.48\%
    \item Evaluation Loss: 0.387
\end{itemize}

The high alignment of metrics indicates robust generalization across imbalanced intent classes.

\subsection{Qualitative Evaluation}
The chatbot was evaluated in real-time dialogues, demonstrating effective intent recognition and response generation:
\begin{itemize}
    \item \textbf{Greeting}: Input: ``wadii'' $\rightarrow$ Intent: Greeting (confidence 1.00) $\rightarrow$ Response: ``Hesi shamwari! Uri sei hako?''
    \item \textbf{Religious Query}: Input: ``mune mufundisi here'' $\rightarrow$ Intent: Religion $\rightarrow$ Response: Department contact link.
    \item \textbf{Finance Query}: Input: ``pace inoita mari?'' $\rightarrow$ Intent: Finance $\rightarrow$ Fallback reply due to limited training data.
    \item \textbf{Application Query}: Input: ``mune ma program api pa Pace'' $\rightarrow$ Intent: Education $\rightarrow$ Interactive dialogue collecting user details.
    \item \textbf{Exit}: Input: ``exit'' $\rightarrow$ Response: ``Zvakanaka, tichaonana zvakare!''
\end{itemize}

\subsection{Baseline Comparison}
The hybrid system was compared to a RAG-only baseline (no intent classification). The hybrid model provided culturally relevant responses (e.g., natural Shona greetings), while the baseline generated generic or off-topic replies, particularly for non-educational queries.

\section{Discussion}
This work addresses the critical need for slang-aware NLP in low-resource African languages \citep{ahia2023bridging, nekoto2020participatory}. The annotated dataset enables robust intent and sentiment detection in code-mixed contexts \citep{eisenstein2013}, while the hybrid architecture balances precision for predictable queries with flexibility for open-ended ones \citep{wang2024rethinking, kloker2024wash}. Key implications include:
\begin{itemize}
    \item \textbf{Enhanced Language Understanding}: Improved handling of informal inputs for social media analysis and user interaction \citep{nguyen2020bertweet}.
    \item \textbf{Culturally Relevant Dialogue}: Context-aware responses that resonate with Shona-speaking users \citep{kepuska2018next}.
    \item \textbf{Digital Inclusion}: Expanded access to AI for low-resource language communities \citep{joshi2020state, ndichu2024african}.
\end{itemize}

\subsection{Use Case: Educational Assistant}
The chatbot was deployed to assist prospective students at Pace University, handling queries in Shona slang and English. It supports greetings, program inquiries, and application workflows, improving accessibility for Shona-speaking users \citep{ahia2023bridging}.

\subsection{Limitations}
The dataset’s small size (~34 utterances) limits coverage, particularly for finance-related intents. Compute constraints on Google Colab restricted hyperparameter tuning. Ethical considerations include ensuring data anonymization and community validation for cultural appropriateness \citep{ndichu2024african, joshi2020state}.

\section{Conclusion and Future Work}
This study presents a Shona--English slang dataset, a fine-tuned DistilBERT classifier, and a hybrid chatbot, available at \url{https://github.com/HappymoreMasoka/Working_with_shona-slang} and \url{https://huggingface.co/HappymoreMasoka}. Achieving 96.4\% accuracy, the system advances African NLP by addressing informal communication \citep{ahia2023bridging, joshi2020state}. Future work includes:
\begin{itemize}
    \item Expanding the dataset to include more intents and languages.
    \item Integrating audio inputs for multimodal dialogue.
    \item Enhancing RAG with domain-adaptive retrieval.
    \item Conducting human-in-the-loop evaluations.
\end{itemize}

By releasing these resources, we contribute to inclusive NLP, bridging the gap between formal models and the dynamic realities of African communication.

\section*{Acknowledgments}

This work was conducted at Pace University under the supervision of Krishna Bathula, Ph.D., in the Seidenberg School of Computer Science and Information Systems. We thank Pace University for providing the necessary resources and support.

\bibliographystyle{plainnat}
\bibliography{custom}

\end{document}